# Probabilistic Knowledge Graph Construction: Compositional and Incremental Approaches

Dongwoo Kim
ANU
Australia
dongwoo.kim@anu.edu.au

Lexing Xie
ANU & Data to Decision CRC
Australia
lexing.xie@anu.edu.au

Cheng Soon Ong
Data61 & ANU
Australia
chengsoon.ong@anu.edu.au

## ABSTRACT

Knowledge graph construction consists of two tasks: extracting information from external resources (knowledge population) and inferring missing information through a statistical analysis on the extracted information (knowledge completion). In many cases, insufficient external resources in the knowledge population hinder the subsequent statistical inference. The gap between these two processes can be reduced by an incremental population approach. We propose a new probabilistic knowledge graph factorisation method that benefits from the path structure of existing knowledge (e.g. syllogism) and enables a common modelling approach to be used for both incremental population and knowledge completion tasks. More specifically, the probabilistic formulation allows us to develop an incremental population algorithm that trades off exploitation-exploration. Experiments on three benchmark datasets show that the balanced exploitation-exploration helps the incremental population, and the additional path structure helps to predict missing information in knowledge completion.

## 1. INTRODUCTION

Relational knowledge graphs formalise our understanding about the world. This in turn helps us reason and infer in a wide range of tasks such as information retrieval, question answering, and semantic parsing [5, 9, 12]. The construction of a knowledge graph is an active research area with many important and challenging research questions. The early stage of knowledge graph construction relies on the **knowledge population** task where the goal is to maximise the number of discovered facts in the form of `(entity1, relation, entity2)` triples. External sources such as Wikipedia are used to extract the triples [8], or human experts encode a prior knowledge manually [2]. Despite efforts towards a comprehensive knowledge graph, even the largest commercial knowledge graph is still far from complete [6]. The **knowledge completion** task has emerged as a complement of the knowledge population task to scale up the knowledge graph construction. Unlike the knowledge population task, the goal of knowledge completion is to correctly predict unknown triples based on a statistical analysis of the known triples [15, 19].

An obstacle to knowledge graph construction is a gap between knowledge population and completion. The more a knowledge graph is populated, the better a statistical model predict the unknown triples. In many cases, however, there is insufficient external resource to extract knowledge, and thus the construction relies solely on the incremental population by human experts, which can be slow and costly. We need an active way of selecting triples to be labelled, in order to maximise the performance of the following knowledge completion. A recent attempt at active incremental population [10] has had difficulties simultaneously achieving a high population and faithful reconstruction.

We propose a statistical relational model, providing a probabilistic framework for both knowledge completion and knowledge population. We formulate bilinear tensor factorisation [19] in a probabilistic way, where entities and relations are embedded into a latent feature space. We propose an extension to the tensor factorisation model that incorporates the path structure of a knowledge graph into the factorisation. The probabilistic formulation provides a natural way of exploiting uncertainty of triples, allowing us to develop an active triple selection for the incremental population. We employ Thompson sampling [22] to find an optimal tradeoff between exploration and exploitation during the active selection.

Based on experiments with three benchmark datasets, we find that the additional path structure helps predict unobserved triples, while the model without the path structure is more helpful in the incremental population. This apparent contradiction in results can be explained by the different requirements on the latent structure. For knowledge completion, it is important to find good latent structures, For incremental population, however, it is more important to accurately estimate uncertainty such that we can explore the latent space efficiently over time. To the best of our knowledge, this is the first study that explicitly investigates the contrasting effects of path structure in knowledge completion and incremental population.

**Related work** The literature on data factorisation and vector space models for relational data is vast. We give a brief overview of related work along three design choices:

**Table 1: The categorisation of factorisation problems with respect to three design considerations (see main text for details).**

| Pr/N-Pr | P/A | M/T/C | References |
|---|---|---|---|
| N-Pr | P | M | [16] |
| N-Pr | A | M | [20] |
| N-Pr | P | T | [19][14] |
| N-Pr | A | T | [10] |
| N-Pr | P | C | [18][7] |
| N-Pr | A | C | – |
| Pr | P | M | [17] |
| Pr | A | M | [11][23] |
| Pr | P | T | *, [25][21] |
| Pr | A | T | * |
| Pr | P | C | * |
| Pr | A | C | * |

the method, the learning strategy, and the data representation. In Table 1, we summarise related work along all combinations in each dimension. Our work address the uncovered combination of three design choices in the probabilistic method, indicated by an asterisk.

**Probabilistic (Pr) / Non-Probabilistic (N-Pr)** This refers to two broad classes of model formulation, whether an obtained model has a probabilistic interpretation.

**Passive (P) / Active (A)** This refers to two different learning strategies, of passively learning a model given labeled data points, or actively requesting data points to be labelled.

**Matrix (M) / Tensor (T) / Composition (C)** This refers to three data representations. Matrix represents single relational data such as `(user, item)`. Tensor represents multiple relational data such as `(entity, relation, entity2)`. Composition includes more complex structures of multiple relational data such as `(entity1 - relation1 - entity2 - relation2 - entity3)`.

Model details and additional results can be found in the online appendix [13].

## 2. PROBABILISTIC RESCAL

A relational knowledge graph consists of a set triples in the form of $(i, k, j)$ where $i$, $j$ are entities, and $k$ is a relation. A triple can be distinguished in a valid triple and invalid triple based on a semantic meaning of the triple. An example of the valid triple in Freebase is `(BarackObama, PresidentOf, U.S.)`, and an example of the invalid triple is `(BarackObama, PresidentOf, U.K.)`. A knowledge graph can be represented in a three-way binary tensor $\mathcal{X} \in \{0,1\}^{N\times K\times N}$, where $K$ is a number of relations, $N$ is a number of entities, and $x_{ikj} \in \{0,1\}$ indicates whether the triple is valid.

We model the entity $i$ as vectors $e_i$ and the relation $k$ as matrix $R_k$ with an appropriately chosen latent dimension $D$. This follows a popular model for statistical relational learning, which is to factorise the tensor into a set of latent vector representations, such as the bilinear model RESCAL [19]. RESCAL aims to factorise each relational slice $\mathcal{X}_{:k:}$ into a set of rank-$D$ latent features as follows:

$$\mathcal{X}_{:k:} \approx E R_k E^\top, \qquad \text{for } k = 1, \ldots, K$$

Here, $E \in \mathbb{R}^{N\times D}$ contains the latent features of the entities $e_1, \ldots, e_N$ and $R_k \in \mathbb{R}^{D\times D}$ models the interaction of the latent features between entities in relation $k$.

We propose a probabilistic framework that directly generalises RESCAL (PRESCAL) by placing priors over the latent features. For each entity $i$, the latent feature of an entity $e_i \in \mathbb{R}^D$ is drawn from an isotropic multivariate-normal distribution with variance $\sigma_e^2$,

$$e_i \sim N(\mathbf{0}, \sigma_e^2 I_D). \tag{1}$$

For each relation $k$, we draw matrix $R_k$ from a zero-mean isotropic matrix normal distribution with variance $\sigma_r^2$,

$$R_k \sim \mathcal{MN}_{D\times D}(\mathbf{0}, \sigma_r I_D, \sigma_r I_D) \tag{2}$$
$$\text{or equivalently } \; r_k = \text{vec}(R_k) \sim N(\mathbf{0}, \sigma_r^2 I_{D^2})$$

where $\text{vec}(R_k)$ denotes the flattening of the matrix.

**PNORMAL** We consider two observation models for $x_{ikj}$: real or binary variables. By placing a normal distribution over $x_{ikj}$,

$$x_{ikj} | e_i, e_j, R_k \sim \mathcal{N}(e_i^\top R_k e_j, \sigma_x^2), \tag{3}$$

we model the value of triple as a real variable. This is not a natural choice since the triple is a binary variable, however, we can control the confidence on different observations through the variance parameter $\sigma_x^2$. We develop a Gibbs sampler to perform the posterior inference for the normally distributed observation model. The conditional distribution of each latent variable is given by:

$$p(e_i | E_{-i}, \mathcal{R}, \mathcal{X}^t, \sigma_e, \sigma_x) = \mathcal{N}(e_i | \mu_i, \Lambda_i^{-1}) \tag{4}$$
$$p(R_k | E, \mathcal{X}, \sigma_r, \sigma_x) = \mathcal{N}(\text{vec}(R_k) | \mu_k, \Lambda_k^{-1}) \tag{5}$$

where the negative subscript $-i$ indicates the every other entity variables except entity $i$. Exact forms of the posterior means and precision matrices are listed in Table 2, where we have used the identity $e_i^\top R_k e_j = r_k^\top e_i \otimes e_j$.

**PLOGIT** One may want to model the binary observation more precisely. Here, we model $x_{ikj}$ as a Bernoulli random variable whose probability is determined by logistic regression:

$$p(x_{ikj} = 1) = \sigma(e_i^\top R_k e_j), \tag{6}$$

where $\sigma$ is a sigmoid function. We approximate the conditional posterior of $E$ and $R$ by the Laplace approximation [1] through an alternative sampling. The detailed derivations are provided in Appendix A.

**Thompson Sampling** The probabilistic framework allows us to quantify the uncertainty of predictive distribution, which is then used to formulate an active learning algorithm. Specifically, we adopt Thompson sampling for active learning, which finds an optimal trade off between exploitation and exploration during active learning. Thompson sampling provides a model based query selection process [3, 22]. Let $x_{1:t}$ be a sequence of observed triples up to time $t$, and $\theta$ is an underlying parameter governing the rewards $r$. Thompson sampling chooses the next action $a$ (triple to label), according to its probability of having high reward:

$$\arg\max_a \int \mathbb{I}\left[\mathbb{E}(r|a,\theta) = \max_{a'} \mathbb{E}(r|a',\theta)\right] p(\theta | x_{1:t-1}) d\theta,$$

where $\mathbb{I}$ is an indicator function. Note that it is sufficient to draw a random sample from the posterior instead of computing the integral.

**Table 2: Parameters for Gibbs updates. The conditional of $e_i$ and $R_k$ follows the normal distribution with mean $\mu$ and precision matrix $\Lambda$. $\otimes$ is the Kronecker product.**

| var | $\mu$ | $\Lambda$ | $\xi$ |
|---|---|---|---|
| $e_i$ | $\frac{1}{\sigma_x^2}\Lambda_i^{-1}\xi_i$ | $\frac{1}{\sigma_x^2}\sum_{jk:x_{ikj}\in\mathcal{X}^t}(R_k e_j)(R_k e_j)^\top$ | $\sum_{jk:x_{ikj}\in\mathcal{X}^t} x_{ikj}R_k e_j + \sum_{jk:x_{jki}\in\mathcal{X}^t} x_{jki}R_k^\top e_j$. |
| $vec(R_k)$ | $\frac{1}{\sigma_x^2}\Lambda_k^{-1}\xi_k$ | $\frac{1}{\sigma_x^2}\sum_{ij:x_{ikj}\in\mathcal{X}^t}(e_i\otimes e_j)(e_i\otimes e_j)^\top + \frac{1}{\sigma_r^2}I_{D^2}$ | $\sum_{ij:x_{ikj}\in\mathcal{X}^t} x_{ikj}(e_i\otimes e_j)$. |

We generalise the idea of a particle Thompson sampling originally proposed in [11] for a matrix factorisation to the tensor factorisation. The detailed algorithm is provided in Appendix B.

# 3. COMPOSITIONAL RELATIONS

In this section, we propose a compositional relation model that exploits the compositional structure of knowledge graphs to capture the latent semantic structure of the entities and relations. A very recent study shows the benefit of using compositionality in the vector space model [7]. Here, we further extend their framework in a probabilistic way.

The compositionality represents a semantic meaning of a path over a knowledge graph that corresponds to a sequence of composable triples. For example, given two triples, "Barack Obama is a 44th president of U.S." (`BarackObama, PresidentOf, U.S`) and "Joe Biden was a running mate of Barack Obama" (`JoeBiden, RunningMateOf, BarackObama`), one can naturally deduce that the "Joe Biden is a vice president of U.S." (`JoeBiden, VicePresidentOf, U.S.`). Here the composition of two relations, president of, and running mate of, yields a compositional relation, vice president of. More formally, if there is a sequence of triples where the target entity of a former triple is a source entity of a latter triple in a consecutive pair of triples in the sequence, then we can form a compositional triple as follows. Given the sequence of $n$ triples $(i_1, k_1, j_1)$, $(i_2, k_2, j_2)$, $(i_3, k_3, j_3)$ $\dots$ $(i_n, k_n, j_n)$, where $j_k = i_{k+1}$ for all $k$, we form a compositional triple $(i_1, c(k_1, k_2, \dots, k_n), j_n)$, where $c$ denotes the compositional relation of the sequence of relations.

Let $\mathcal{C}^L$ be a set of all possible compositions whose length is up to $L$, $c \in \mathcal{C}$ be a sequence of relations, $c(i)$ be $i$th index of a relation in sequence $c$ and $|c|$ be the length of the sequence. With set of compositions $\mathcal{C}^L$, we can expand set of observed triples $\mathcal{X}$ to set of compositional triples $\mathcal{X}^{\mathcal{C}^L}$ in which compositional triple $x_{icj}$ is an indicator variable that show the existence of the path from entity $i$ to entity $j$ through sequence of relations $c$ in $\mathcal{X}$. Note that the compositional relation $c$ is an abstract relation, and there might be a multiple possible paths from entity $i$ to $j$.

With these extended compositional triples, we again model $x_{icj}$ with a bilinear Gaussian distribution,

$$x_{(i,c(k_1,k_2,\dots,k_n),j)} \sim \mathcal{N}(e_i^\top R_{c(k_1,k_2,\dots,k_n)} e_j, \sigma_c^2), \qquad (7)$$

where $R_{c(k_1,k_2)} \in \mathbb{R}^{D\times D}$ is a latent matrix of compositional relation $c$, and $\sigma_c^2$ is a covariance of the compositional triples. Again the entity vectors are shared across the compositional and non-compositional triples. With the compositions of relations, the PRESCAL may place a new relation matrix $R_c$ for each composition $c$. However the number of required matrices increases exponentially with respect to the length of composition. Consequently, the computational cost will also increase exponentially. To limit the required number of parameters, we propose two different ways of modelling the compositional relation $R_c$.

## 3.1 Additive Compositionality

We define an additive compositional relation $R_c$ as a normalised sum over the sequence of relation matrices in composition $c$, i.e., $R_c = \frac{1}{|c|}(R_{c(1)} + R_{c(2)} + \dots + R_{c(|c|)})$, then compositional triple $x_{icj}$ is modelled as

$$\begin{aligned} x_{(i,c,j)} &\sim \mathcal{N}(e_i^\top R_c e_j, \sigma_c^2) \qquad (8)\\ &= \mathcal{N}(e_i^\top \frac{1}{|c|}(R_{c(1)} + R_{c(2)} + \dots + R_{c(|c|)})e_j, \sigma_c^2). \end{aligned}$$

This treats the relations as vectors, finding the average of a sequence of composed relations. The conditional distribution of $e_i$ and $R_k$ given the rest can be obtained in the same way used for the posterior distribution of PRESCAL. Parameter estimation is shown in Appendix C.1.

## 3.2 Multiplicative Compositionality

Second, we define an multiplicative compositional relation $R_c$ as a sequence of multiplication over relations in composition $c$, i.e. $R_c = R_{c(1)}R_{c(2)}\dots R_{c(|c|)}$, and the compositional triple as a bilinear Gaussian distribution with the compositional relation $R_c$,

$$x_{(i,c,j)} \sim \mathcal{N}(e_i^\top R_{c(1)}R_{c(2)}\dots R_{c(|c|-1)}R_{c(|c|)}e_j, \sigma_c^2) \qquad (9)$$

The multiplicative compositionality can be understood as a sequence of linear transformation from the original entity $i$ with the compositional relations, and the inner product between the transformed entity and target entity forms a value of the compositional triple. Again, the details of the parameter estimators are shown in Appendix C.2.

In contrast with the additive model, the multiplicative model preserves the ordering in a compositional relation, and hence the different orderings of relations result different relations in the compositional model.

# 4. KNOWLEDGE COMPLETION

We first evaluate our model for the knowledge completion task to measure the predictive performance of PRESCAL with all non compositional and compositional variants. We evaluate the models on three benchmark datasets: KINSHIP, UMLS, and NATION, and compare performances with the original RESCAL. Detailed description of each dataset is shown in Table 3.

**Table 3: Description of datasets. Sparsity denotes the ratio of valid triples to invalid triples.**

| Dataset | # rel | # entities | # triples | sparsity |
|---|---|---|---|---|
| Kinship | 26 | 104 | 10,790 | 0.038 |
| UMLS | 49 | 135 | 6,752 | 0.008 |
| Nation | 56 | 14 | 2,024 | 0.184 |

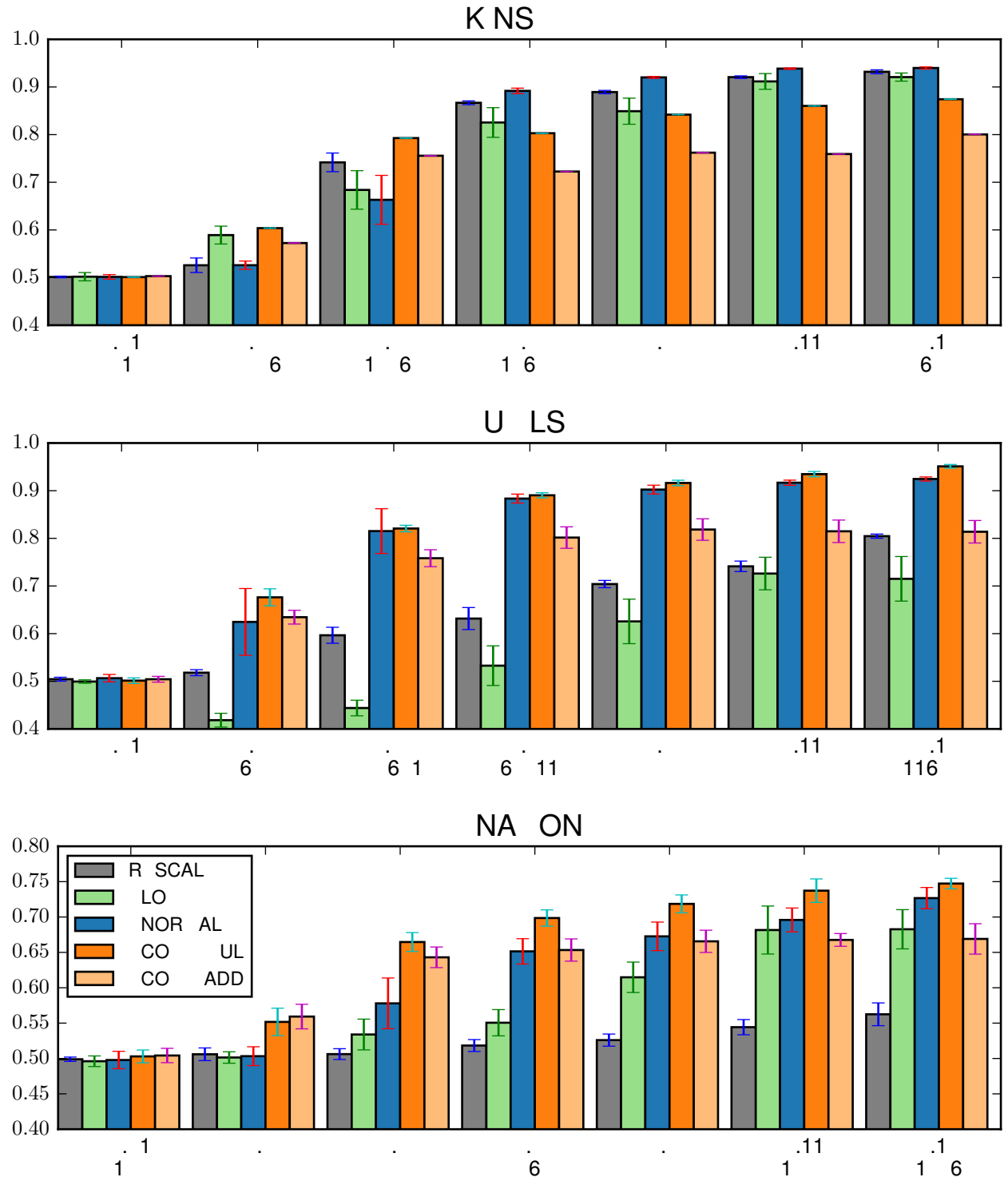

**Figure 1: ROC-AUC scores of compositional models. The x-axis denotes the proportion and total number of triples used for training. Error bars denote one standard deviation.**

We set the compositional length $L$ to two, split the dataset into 20% for validation and 30% for testing. We vary the proportion of training triples from 1% to 13% of datasets. For Rescal, we use the original implementation[1], and measure performance over 10 runs with random initialisations. For Prescal and all the variants, we sample triples $x_{ikj}$ from its posterior, and measure performance over 10 different samples. The performances of models are measured by the ROC-AUC score on the test set ranked according to a posterior mean: $\frac{1}{|\mathcal{X}_p||\mathcal{X}_n|}\sum_{\{i,k,j\}\in\mathcal{X}_p,\{i',k',j'\}\in\mathcal{X}_n}\mathbb{I}[\bar{x}_{ikj} > \bar{x}_{ikj}]$, where $\mathcal{X}_p$ and $\mathcal{X}_n$ are the set of positive and negative triples in the test set, respectively, and $\bar{x}$ is a reconstructed triple.

Figure 1 shows the ROC-AUC scores of the compositional models with the various baseline models. The Prescal with the normal output (Pnormal) or logistic output (Plogit) generally outperform Rescal, and Pnormal outperforms Plogit. We conjecture that an additional flexibility of controlling the variance of triples makes Pnormal to perform better than Plogit. We compare the compositional model with the original Rescal, Pnormal, and Plogit. In general, the multiplicative compositional model (Pcomp-mul) outperforms the additive compositional model (Pcomp-add), and performs better than the other baseline models when the training set is small. For UMLS and NATION, Pcomp-mul has the best performance across the all training proportions. For KINSHIP, however, Pcomp-mul performs better when the training proportion is less than 7%.

The goal of the compositional models is to factorise triples along with the graph structure as a whole. The triple prediction task tells us a trained model is capable of triple prediction, but does not tell whether the features can recover the graph structure. If the model factorises the graph structure properly, then the trained model can predict not only triples but the graph structure as well. To validate the model assumption, we evaluate a path prediction task. For this task, we use 10% of UMLS dataset for training. We compute an expected value of an unobserved path given a trained model. The non-compositional models may not be able to compute the expected value. In such a case, we approximate paths with the multiplicative model assumption in Equation 9. We vary the path length from 1 (triple) to 4, and measure ROC-AUC scores on the reconstructed compositional triples. Figure 2 shows the result of the path prediction task. Both compositional models show consistent performance regardless of the path length. However, the performance of the non-compositional models drops sharply as the length increases. The results show the compositional models preserve graph structure in the embedded space. It is worth emphasising that although the compositional length for training is 2, the compositional models show consistent results on predicting paths of length 3 and 4.

[1] https://github.com/mnick/rescal.py

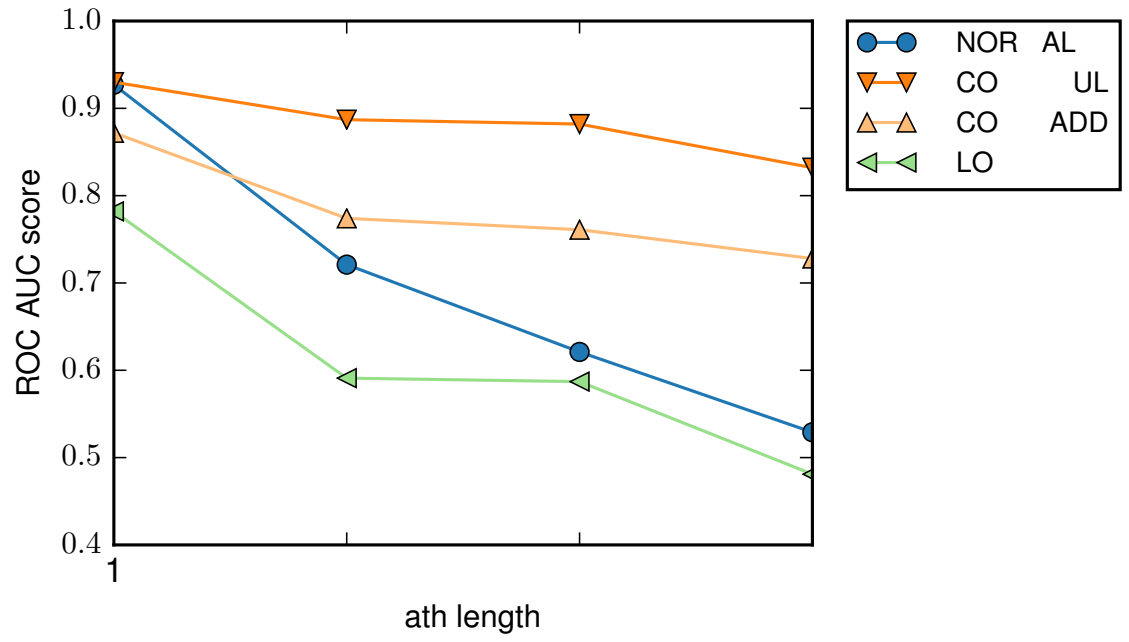

**Figure 2: Path prediction result with UMLS. The performances of both compositional models remain consistent whereas those of the non-compositional models drop sharply as the length increases.**

Table 4 shows an example of the path prediction result starting from entity `Mental-or-Behavioral (MB) Dysfunction` followed by two relations `Affects` and `Produces` in the UMLS dataset. Both compositional and non-compositional models predict triples well. For length-2 path prediction, only the compositional models can capture correct entities on top 5. We also visualise the multi-dimensional entities inferred by Pnormal and Pcomp-mul into a two-dimensional space using spectral clustering [24] in Appendix D.

# 5. KNOWLEDGE POPULATION

In this section, we show results of incremental knowledge population task using Thomson sampling on the three datasets. Additional verification on the Thompson sampling with synthetic datasets is also provided in Appendix E.

**Experimental settings**: We compare the Thompson sampling models with Amdc models, and Prescal for passive learning. Amdc model has been proposed to achieve two different active learning goals, constructing a predictive model and maximising the valid triples in a knowledge base, with two different querying strategies [10]. Amdc-pred

**Table 4: Example of path prediction from UMLS data. We predict top 5 entities in compositional triples starting from entity `Mental-or-Behavioral (MB) Dysfunction` followed by two relations `Affects` and `Produces`. Correct entities are bolded.**

(a) Triple prediction: `(MB Dysfunction, Affects, -)`

| Model | Top 1 | Top 2 | Top 3 | Top 4 | Top 5 |
|---|---|---|---|---|---|
| PNORMAL | **Invertebrate** | **Reptile** | **Archaeon** | **Bird** | **Phy.-Function** |
| PLOGIT | **Cell-Function** | **Disease-or-Syndrome** | **Cell-or-Molecular-Dysf.** | **Exp.-Model-of-Disease** | **Mental-Process** |
| PCOMP-MUL | **Archaeon** | **Fish** | **Fungus** | **Invertebrate** | **Human** |
| PCOMP-ADD | **Path.-Function** | **Bird** | **Cell-or-Molecular-Dysf.** | Drug-Delivery-Device | Congenital-Abnormality |

(b) Length-2 path prediction: `(MB Dysfunction, Affects, Produces, -)`

| Model | Top 1 | Top 2 | Top 3 | Top 4 | Top 5 |
|---|---|---|---|---|---|
| PNORMAL | Clinical-Drug | Sign-or-Symptom | Org.-Attribute | Drug-Delivery-Device | Clinical-Attr. |
| PLOGIT | Amphibian | Gov.-or-Reg.-Activity | Food | Biologic-Func. | Classification |
| PCOMP-MUL | **Enzyme** | **Body-Substance** | **Biogenic-Amine** | Carbohydrate | **Immunologic-Factor** |
| PCOMP-ADD | **Immunologic-Factor** | **Body-Substance** | Molecular-Biology-Research-Technique | Clinical-Drug | Chemical-Viewed-Structurally |

is a predictive model construction strategy and chooses a triple which is the most ambiguous (close to the decision boundary) at each time $t$. AMDC-POP is a population strategy which aims to maximise the number of valid triples in a knowledge base, choosing a triple with the highest expected value at each time. To train all models we only use the observed triples up to the current time. For the passive learning with PRESCAL, we generate a random sample at each time period. For the particle Thompson sampling models, we set variance parameter $\sigma_e$ and $\sigma_r$ to 1, $\sigma_x$ to 0.1, and vary $\sigma_c$ from 1 to 100.

We leave 30 % of triples as a test set to measure test error. At each time period, each model chooses one triple to query, if the selected triple is in the test set then we choose the next highest expected triple that is not in the test set. All models start from zero observation. After every query, a model obtains a label of the queried triple from an oracle, then the model updates the parameters.

**Evaluation metric**: We use two different evaluation metrics, the cumulative gain and ROC-AUC score, for the performance comparison. The goal of the Thompson sampling is to maximise the knowledge population through the balanced querying strategy between exploration and exploitation. To measure how many triples are obtained through the querying stage, we compute the cumulative gain which is the number of valid triple obtained up to time $t$. Additionally, we compute the ROC-AUC score on the test set to understand how this balanced querying strategy results in making a predictive model.

**Exploitation and exploration**: Figure 3 shows the cumulative gains and ROC-AUC scores of the Thompson sampling on three real datasets. The model names with suffix -TS represent the models adopting the Thompson sampling strategy. PNORMAL-TS performs better than other baseline models for the cumulative gain, and shows comparable result for the ROC-AUC scores. Both compositional models perform worse than PNORMAL-TS across all datasets.

In the original AMDC [10], AMDC-POP model obtains more valid triples than AMDC-PRED, and AMDC-PRED shows high ROC-AUC scores than AMDC-POP. In our experiment, however, AMDC-POP shows comparable cumulative gain to AMDC-PRED and even worse than AMDC-PRED for the UMLS. We conjecture the initial observation and query size results in the different performances: in the original experiment, the model starts from a small set of training data, and the query size was 1,000 for KINSHIP and UMLS. With larger query size, the model focuses on exploit and takes advantages, whereas in our experiment, we start from zero observation and query one triple at each time, which makes the model hard to exploit. This result shows the importance of balancing between exploitation and exploration.

We note that the compositional model performs worse than the non-compositional models, especially than PNORMAL-TS. This is counter-intuitive to our general understanding where the model that performs well in the predictive task also shows a better performance in the active learning.

# 6. CONCLUSION

Throughout the paper, we have considered the two knowledge base construction tasks: knowledge population and knowledge completion. Based on a probabilistic framework, we propose new knowledge base factorisation methods where the latent factorisation reflects the graph structure of a knowledge graph. The probabilistic formulation allows us to quantify the uncertainty of predictive distributions, which is then used for the knowledge population task. The experiments of two tasks on three datasets show that the compositional model benefits graph structure for knowledge completion, and the probabilistic formulation helps to explore the latent space efficiently for knowledge population.

**Acknowledgments** This work is supported in part by the Australian Research Council via the Discovery Project program DP140102185. The work has been supported by the Data to Decisions Cooperative Research Centre whose activities are funded by the Australian Commonwealth Government's Cooperative Research Centres Programme.

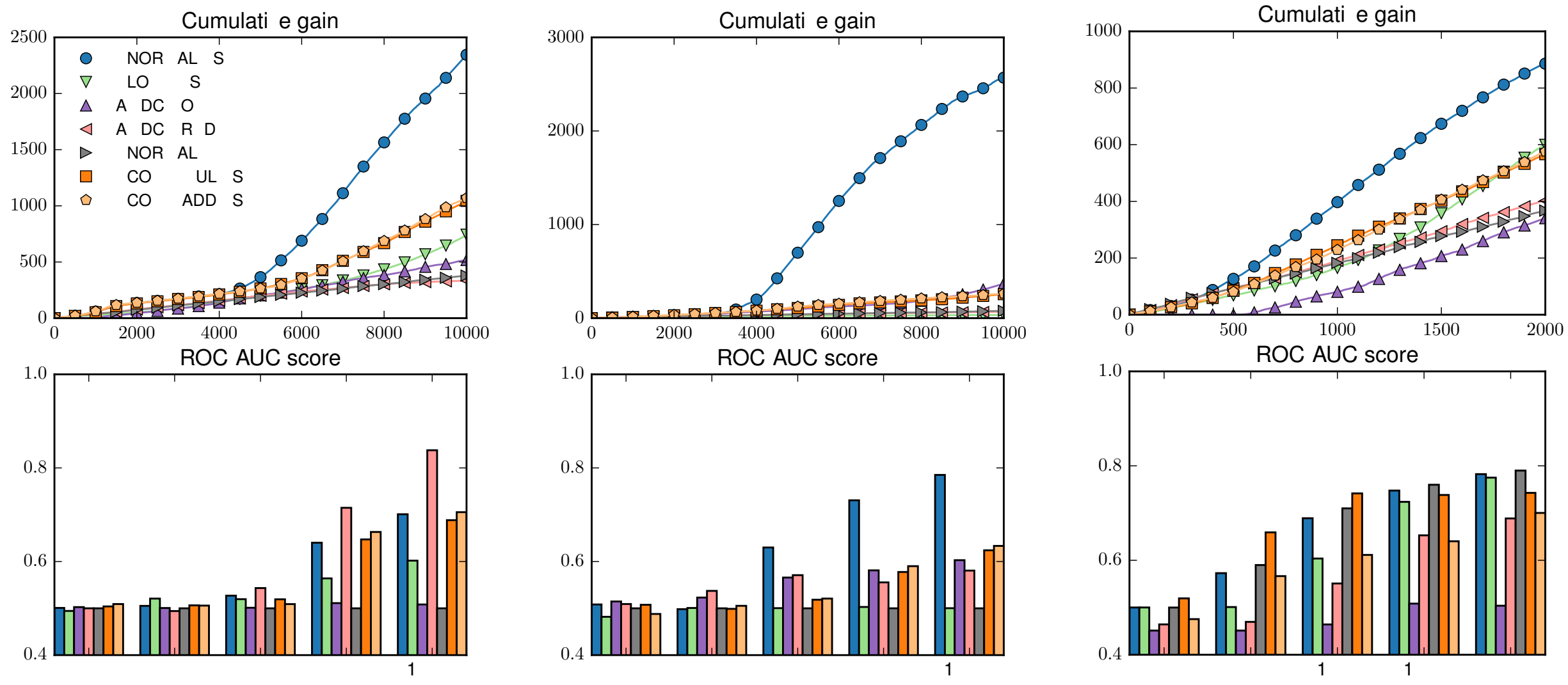

(a) KINSHIP (b) UMLS (c) NATION

**Figure 3: The cumulative gain and ROC-AUC score of the Thompson sampling with passive learning and AMDC models. Thompson sampling with PRESCAL (PNORMAL-TS) model achieves the highest cumulative gain to compare with the other models and shows comparable performance on ROC-AUC scores.**

# APPENDIX

# A. POSTERIOR OF LOGISTIC OUTPUT

In this section, we provide the posterior distribution of PLOGIT. The maximum a posterior estimate of $e_i$ or $R_k$ given the rest can be computed through standard logistic regression solvers with the priors over $e_i$ and $R_k$ as regularisation parameters. Given the maximum a posteriori parameters $e_i^*$, the posterior covariance $S_i$ of entity $i$ takes the form

$$S_i^{-1} = \sum_{x_{ikj}} \sigma(e_i^{*\top} R_k e_j)(1 - \sigma(e_i^{*\top} R_k e_j)) R_k e_j (R_k e_j)^\top$$

$$+ \sum_{x_{jki}} \sigma(e_j^\top R_k e_i^*)(1 - \sigma(e_j^\top R_k e_i^*)) R_k^\top e_i^* (R_k^\top e_i^*)^\top + I\sigma_e^{-1}.$$

The posterior covariance of $R_k$ can be computed in the same way. Let $R_k^*$ is a maximum a posterior solution of $R_k$ given $E$. Then, the conditional posterior covariance $S_k$ of relation $k$ has the form of:

$$S_k^{-1} = \sum_{x_{ikj}} \sigma(e_i^\top R_k^* e_j)(1 - \sigma(e_i^\top R_k^* e_j)) \bar{e}_{ij} \bar{e}_{ij}^\top + I\sigma_r^{-1},$$

where $\bar{e}_{ij} = e_i \otimes e_j$.

# B. PARTICLE THOMPSON SAMPLING

We introduce a particle Thompson sampling algorithm for the incremental knowledge population with the proposed PRESCAL models. In the incremental population task, a knowledge base starts with some initial observations, and at each time period, the agent select one triple to be labelled by an external system, i.e. human experts. The queried triple is chosen selectively based on past observations. Each label incurs a cost, so the goal is to obtain as many valid triples as possible within a given budget.

Thompson sampling provides a model based query selection process, and has been gaining an increasing attention because of its competitive empirical performance as well as conceptual simplicity [3, 22]. Let $y_{1:t}$ be a sequence of rewards up to time $t$, and $\theta$ is an underlying parameter governing the rewards. With Thompson sampling, an agent chooses action $a$ according to its probability of being optimal:

$$\arg\max_a \int \mathbb{I}\Big[\mathbb{E}(r|a,\theta) = \max_{a'} \mathbb{E}(r|a',\theta)\Big] p(\theta|y_{1:t-1}) d\theta,$$

where $\mathbb{I}$ is an indicator function. Note that it is sufficient to draw a random sample from the posterior instead of computing the integral.

We formulate Thompson sampling for incremental knowledge population system as follows. First, we assume there are optimal latent features $E^*$ and $R^*$, and the triples are generated through Equation 1− 3. At time $t$, the system draws samples $E^t$ and $R^t$ from the posterior distribution, and then chooses an optimal triple $(i,k,j)^* = \arg\max_{i,k,j} e_i^\top R_k e_j$ to be queried. Finally, with the newly observed triple $x_{(i,k,j)^*}$, the system updates the posterior of the latent features.

The main difficulty of applying Thompson sampling is a sequential update for the posterior of the latent features with new observations over time. Unlike the point estimation algorithms such as the maximum likelihood estimator, computing a full posterior with MCMC requires extensive computational cost. To make the algorithm feasible, we employ a sequential Monte-Carlo (SMC) method for online posterior inference, generalising an algorithm proposed in [11] to tensors.

**Algorithm 1** Particle Thompson sampling for probabilistic RESCAL with Gaussian output variable

**Input:** $\mathcal{X}^0, \sigma_x, \sigma_e, \sigma_r$.
**for** $t = 1, 2, \ldots$ **do**
  *Thompson Sampling*:
  $h_t \sim \mathrm{Cat}(\mathbf{w}^{t-1})$
  $(i,k,j) \leftarrow \arg\max p(x_{ikj}|E^{h_{t-1}}, \mathcal{R}^{h_{t-1}})$
  Query $(i,k,j)$ and observe $x_{ikj}$
  $\mathcal{X}^t \leftarrow \mathcal{X}^{t-1} \cup \{x_{ikj}\}$
  *Particle Filtering*:
  $\forall h, w_h^t \propto p(x_{ikj}|E^h, \mathcal{R}^h)$ ▷ Reweighing
  **if** $\mathrm{ESS}(\mathbf{w}^t) \le N$ **then**
    resample particles
    $w_h^t \leftarrow 1/H$
  **end if**
  **for** $h = 1$ **to** $H$ **do**
    $\forall k, R_k^{h_t} \sim p(R_k|\mathcal{X}^t, E^{h_{t-1}}, \mathcal{R}_{-k})$ ▷ see Table (2)
    $\forall i, e_i^{h_t} \sim p(e_i|\mathcal{X}^t, E_{-i}, \mathcal{R}^{h_t})$ ▷ see Table (2)
  **end for**
**end for**

The SMC starts with $H$ number of particles, each of which starts with likelihood weight $w_h = 1/H$, and a set of randomly sampled latent features $E^{h_0}$ and $\mathcal{R}^{h_0}$. With a slight abuse of notation, let $\mathcal{X}^t$ be a set of observed triples up to time $t$. At time $t$, the system chooses one particle according to the particle weights, and then generates a new query via Thompson sampling from the selected particle. After observing a new variable, the system updates the posterior samples of every particle through the MCMC kernels with the new observation. We first sample the relation matrices using Equation 5, and sample the entity vectors using Equation 4. Under the mild assumption where $p(\Theta|\mathcal{X}^{t-1}) \approx p(\Theta|\mathcal{X}^t)$, $\Theta = \{E, \mathcal{R}\}$, the weight of each particle at time $t$ can be computed as follows [4]:

$$w_h^t = \frac{p(\mathcal{X}^t|\Theta)}{p(\mathcal{X}^{t-1}|\Theta)} = p(x^t|\Theta, \mathcal{X}^{t-1}) \tag{10}$$

To keep the posterior samples on regions of high probability mass, we resample the particles whenever an effective sample size (ESS) is less than a predefined threshold. The ESS can be computed as $(\sum_h w_h^2)^{-1}$, and we set the threshold to $N/2$. Resampling removes low weight particles with high probability, while keeping samples from the posterior. We summarise the particle Thompson sampling for PRESCAL with the Gaussian output variable in Algorithm 1[2].

Both compositional models can use the same particle Thompson sampling scheme described in Algorithm 1 with the conditional distributions. However, the model can only query the triples in the original tensor and not in the expanded tensor because the compositional triples are unobservable.

We show that the Thompson sampling approach improves over passive PRESCAL in experiments with real and synthetic data. We also investigated the extension of the Rao-Blackwellisation approach as proposed in [11], but we did not observe any significant performance improvements.

[2] Download code here: https://git.io/vi3ZQ.

## C. POSTERIOR DISTRIBUTION OF COMPOSITIONAL RELATIONS

We provide the conditional posterior distributions of two compositional models.

### C.1 Additive Compositionality

The conditional distribution of $e_i$ given $E_{-i}, \mathcal{R}, \mathcal{X}^t, \mathcal{X}^{L(t)}$ is expanded from the posterior of PRESCAL by incorporating compositional triples.

$$p(e_i|E_{-i}, \mathcal{R}, \mathcal{X}^t, \mathcal{X}^{L(t)}) = \mathcal{N}(e_i|\mu_i, \Lambda_i^{-1}), \quad (11)$$

where

$$\mu_i = \Lambda_i^{-1}\xi_i$$
$$\Lambda_i = \frac{1}{\sigma_x^2}\sum_{jk:x_{ikj}\in\mathcal{X}^t}(R_k e_j)(R_k e_j)^\top + \frac{1}{\sigma_x^2}\sum_{jk:x_{jki}\in\mathcal{X}^t}(R_k^\top e_j)(R_k^\top e_j)^\top + \frac{1}{\sigma_c^2}\sum_{jc:x_{icj}\in\mathcal{X}^{L(t)}}(R_c e_j)(R_c e_j)^\top + \frac{1}{\sigma_c^2}\sum_{jc:x_{jci}\in\mathcal{X}^{L(t)}}(R_c^\top e_j)(R_c^\top e_j)^\top + \frac{1}{\sigma_e^2}I_D$$
$$\xi_i = \frac{1}{\sigma_x^2}\sum_{jk:x_{ikj}\in\mathcal{X}^t} x_{ikj}R_k e_j + \frac{1}{\sigma_x^2}\sum_{jk:x_{jki}\in\mathcal{X}^t} x_{jki}R_k^\top e_j + \frac{1}{\sigma_c^2}\sum_{jc:x_{icj}\in\mathcal{X}^{L(t)}} x_{icj}R_c e_j + \frac{1}{\sigma_c^2}\sum_{jc:x_{jci}\in\mathcal{X}^{L(t)}} x_{jci}R_c^\top e_j$$

To compute the conditional distribution of $R_k$, we first decompose $R_c$ into two part where $R_c = \frac{1}{|c|}R_k + \frac{|c|-1}{|c|}R_{c/k}$, where $R_{c/k} = \sum_{k'\in c/k} R_{k'}$. The distribution of compositional triple is decomposed as follows:

$$x_{(i,c,l)} \sim \mathcal{N}(e_i^\top(\frac{1}{|c|}R_k + \frac{|c|-1}{|c|}R_{c/k})e_j, \sigma_c^2). \quad (12)$$

Then, the conditional distribution $R_k$ given $R_{-k}, E, \mathcal{X}^t, \mathcal{X}^{L(t)}$ is

$$p(R_k|E, \mathcal{X}^t, \mathcal{X}^{L(t)}, \sigma_r, \sigma_x) = \mathcal{N}(\mathrm{vec}(R_k)|\mu_k, \Lambda_k^{-1}), \quad (13)$$

where

$$\mu_k = \Lambda_k^{-1}\xi_k$$
$$\Lambda_k = \frac{1}{\sigma_x^2}\sum_{ij:x_{ikj}\in\mathcal{X}^t}\bar{e}_{ij}\bar{e}_{ij}^\top + \frac{1}{\sigma_r^2}I_{D^2} + \frac{1}{|c|^2\sigma_c^2}\sum_{ij:x_{icj}\in\mathcal{X}^{L(t)},\ k\in c}\bar{e}_{ij}\bar{e}_{ij}^\top$$
$$\xi_k = \frac{1}{\sigma_x^2}\sum_{ij:x_{ikj}\in\mathcal{X}^t} x_{ikj}\bar{e}_{ij} + \frac{1}{|c|\sigma_c^2}\sum_{ij:x_{icj}\in\mathcal{X}^{L(t)},\ k\in c} x_{icj}\bar{e}_{ij} - \frac{|c|-1}{|c|}\bar{e}_{ij}r_{c/k}^\top\bar{e}_{ij}$$
$$\bar{e}_{ij} = e_i \otimes e_j.$$

Vectorisation of $R_c$ and $R_{c/k}$ are represented as $r_c$ and $r_{c/k}$, respectively.

The detail derivation of the posterior distribution is as follows:

$$p(R_k|E, R_{-k}, \mathcal{X}) \propto p(\mathcal{X}|R, E)p(R_k)$$
$$\propto \prod_{x_{ikj}} \exp\left\{-\frac{(x_{ikj} - e_i^\top R_k e_j)^2}{2\sigma_x^2}\right\} \prod_{x_{icj}} \exp\left\{-\frac{(x_{icj} - e_i^\top R_c e_j)^2}{2\sigma_c^2}\right\} \exp\left\{-\frac{r_k^\top r_k}{2\sigma_r^2}\right\}$$
$$= \exp\left\{-\frac{\sum_{x_{ikj}}(x_{ikj} - \bar{e}_{ij}^\top r_k)^2}{2\sigma_x^2} - \frac{\sum_{x_{icj}}(x_{icj} - \bar{e}_{ij}^\top r_c)^2}{2\sigma_c^2} - \frac{r_k^\top r_k}{2\sigma_r^2}\right\}$$
$$= \exp\left\{-\frac{\sum_{x_{ikj}}(x_{ikj} - \bar{e}_{ij}^\top r_k)^2}{2\sigma_x^2} - \frac{\sum_{x_{icj}}(x_{icj} - \bar{e}_{ij}^\top(\frac{1}{|c|}r_k + \frac{|c|-1}{|c|})r_{c/k})^2}{2\sigma_c^2} - \frac{r_k^\top r_k}{2\sigma_r^2}\right\}$$
$$= \exp\left\{-\frac{\sum_{x_{ikj}} -2x_{ikj}\bar{e}_{ij}^\top r_k + r_k^\top\bar{e}_{ij}\bar{e}_{ij}^\top r_k}{2\sigma_x^2} - \frac{\sum_{x_{icj}}\frac{2}{|c|}r_k^\top(x_{icj} - \frac{(|c|-1)}{|c|}\bar{e}_{ij}^\top r_{c/k}) + (\frac{1}{|c|^2}r_k^\top\bar{e}_{ij}\bar{e}_{ij}^\top r_k)}{2\sigma_c^2} - \frac{r_k^\top r_k}{2\sigma_r^2} + const\right\}$$
$$\propto \exp\left\{-\frac{1}{2}r_k^\top(\frac{1}{\sigma_x^2}\sum_{x_{ikj}}\bar{e}_{ij}\bar{e}_{ij}^\top + \frac{1}{|c|^2\sigma_c^2}\sum_{x_{icj}}\bar{e}_{ij}\bar{e}_{ij}^\top + \frac{1}{\sigma_r^2}I)r_k - r_k^\top\Big(\frac{\sum_{x_{ikj}} -x_{ikj}\bar{e}_{ij}}{\sigma_x^2} + \frac{\sum_{x_{icj}}|c|^{-1}(x_{icj} - \frac{(|c|-1)}{|c|}\bar{e}_{ij}^\top r_{c/k})}{\sigma_c^2}\Big)\right.$$

Completing the square results Equation 13.

### C.2 Multiplicative Compositionality

Given a sequence of relations including relation $k$, $R_k$ is placed in the middle of the compositional sequence, i.e., $e_i^\top R_{c(1)}R_{c(2)}\dots R_{c(\delta_k)}\dots R_{c(|c|-1)}R_{c(|c|)}e_j$, where $\delta_k$ is the index of relation $k$. For notational simplicity, we will denote the left side $e_i^\top R_{c(1)}R_{c(2)}\dots R_{c(\delta_k-1)}$ as $\bar{e}_{ic(:\delta_k)}^\top$, and the right side $R_{c(\delta_k+1)}\dots R_{c(|c|-1)}R_{c(|c|)}e_j$ as $\bar{e}_{ic(\delta_k:)}$, therefore we can rewrite the mean parameter as $\bar{e}_{ic(:\delta_k)}^\top R_k\bar{e}_{ic(\delta_k:)}$. With the simplified notations, the conditional of $R_k$ is

$$p(R_k|E, \mathcal{X}, \sigma_r, \sigma_x) = \mathcal{N}(\mathrm{vec}(R_k)|\mu_k, \Lambda_k^{-1}), \quad (14)$$

where

$$\mu_k = \Lambda_k^{-1}\xi_k$$
$$\Lambda_k = \frac{1}{\sigma_x^2}\sum_{ij:x_{ikj}\in\mathcal{X}^t}(e_i\otimes e_j)(e_i\otimes e_j)^\top + \frac{1}{\sigma_r^2}I_{D^2} + \frac{1}{\sigma_c^2}\sum_{ij:x_{icj}\in\mathcal{X}^{L(t)},\ k\in c}(\bar{e}_{ic(:\delta_k)}\otimes\bar{e}_{jc(\delta_k:)})(\bar{e}_{ic(:\delta_k)}\otimes\bar{e}_{jc(\delta_k:)})^\top$$
$$\xi_k = \frac{1}{\sigma_x^2}\sum_{ij:x_{ikj}\in\mathcal{X}^t} x_{ikj}(e_j\otimes e_i) + \frac{1}{\sigma_c^2}\sum_{ij:x_{icj}\in\mathcal{X}^{L(t)},\ k\in c} x_{icj}(\bar{e}_{ic(:\delta_k)}\otimes\bar{e}_{jc(\delta_k:)}).$$

The conditional distribution of $e_i$ given the rest is the same as the additive compositional case.

# D. VISUALISATION

In Figure 4, we visualise the multi-dimensional entities inferred by PNORMAL and PCOMP-MUL into a two-dimensional space through the spectral clustering [24]. A circle represents an entity, and the size of the circle is proportional to the uncertainty of the entity in the latent space. In the UMLS dataset, the entities are categorised into 15 types, e.g. `Disorders`, `Living-Beings`, `Phenomena`, etc. We use the same color to represent the entities with the same type. The entities with the same type are located closer to each other with PCOMP-MUL than PNORMAL.

# E. THOMPSON SAMPLING ON SYNTHETIC DATA

In this section, we verify the sequential Thompson sampling through a compositional and non-compositional synthetic data sets.

## E.1 Thompson sampling on non-compositional synthetic data

We first synthesise two datasets following the model assumptions in Section 2. First, entities and relations are generated from zero-mean isotropic multivariate normal distribution, with variance parameters $\sigma_e = 1$, $\sigma_r = 1$ (Eqn. 1 to 2), respectively. We generate two sets of output triples, with the logistic output (Eqn. 6) and the Gaussian with $\sigma_x$ set to 0.1 (Eqn. 3), respectively .

To measure performance, we compute cumulative regret at each time $n$ as $R(n) = \sum_{t=1}^{n} x_t - x_t^*$, where $x_t^*$ is the highest-valued triple among triples that have not been chosen up to time $t$. Unlike the general bandit setting where one can select a single item multiple times, in our formulation, we can select one triple only once. So after selecting a triple at time $t$, the selected triple will be removed from a set of candidate triples.

Figure 5 shows the cumulative regret of the algorithm on the synthetic data with varying size of entities and relations. We compare the cumulative regret of the particle Thompson sampling with the passive learning method where the model choose a random triple at each time. All results are averaged over 10 individual runs with different initialisations. Note that the dataset with binary logistic output variables can be used to train both logistic-output PRESCAL (PLOGIT) and Gaussian-output PRESCAL (PNORMAL) whereas the dataset with the Gaussian output can only be trained by PNORMAL. Figure 5(a) and 5(b) show that with the logistic synthetic dataset both models are capable to learn the latent features of the generated triples, with logistic outperforming the Gaussian; Figure 5(c) and 5(d) show that the Thompson sampling for PNORMAL (PNORMAL-TS) outperform the passive learning in the real valued dataset.

## E.2 Thompson sampling on compositional synthetic data

We conduct a second experiment on synthetic dataset to understand how the Thompson sampling works for the compositional data. As in the first experiment, we first generate entities and relations from zero-mean multivariate normal with variance parameter $\sigma_e = 1$ and $\sigma_r = 1$. We generate a set of triples with Gaussian output as in Equation 3. We then synthesise two sets of expanded tensors using the previously used entities and relations based on the multiplicative and additive compositional assumptions, defined in Sec 3, respectively. So we synthesise fully observable expanded tensor $\mathcal{X}^L$ where $L = 2$. We set both variance parameter $\sigma_x$ and $\sigma_c$ to 0.1. Note that in a real world situation, the expanded tensor can only be constructed through the observed triples, and the triples in the expanded tensor cannot be queried.

To run the particle Thompson sampling on the synthetic dataset, we let the compositional models know which relation is composed by other relations. The non-compositional PNORMAL model assumes each relation is independent to one another. Therefore, the compositional model uses much less number of parameters to model the same size of tensor to compare with the non-compositional model. With this fully observable expanded tensors, we run the Thompson sampling of the compositional models. Figure 6 shows the cumulative regrets on synthetic datasets. The multiplicative and additive compositionality are used to generate the dataset for Figure 6(a) and 6(b), respectively. The results correspond to our assumption: the Thompson sampling for multiplicative compositional model (PCOMP-ADD-TS) shows lower regrets on the multiplicative data in Figure 6(a), and the Thompson sampling for additive compositional model (PCOMP-ADD-TS) shows lower regrets on the additive compositional data in Figure 6(b), and both have lower regrets than passive learning or PNORMAL-TS without compositions.

# F. POSTERIOR VARIANCE ANALYSIS

In section 5, we find that the compositional model performs worse than the non-compositional models in the active incremental population. We emphasise the difference between two experiments; the goal of incremental population is to maximise the number of triples whereas the goal of knowledge completion in Section 4 is to maximise the predictive performance. Nevertheless, the compositional models do not outperform PNORMAL-TS in the active learning. This result can be partially understood in terms of the balance between exploration-exploitation. Figure 7 shows the average posterior variance of the entity vectors. We compute the eigenvalues of posterior covariance matrix $\Lambda_i^{-1}$ and trace the average eigenvalues over the iterations. As shown in the figure, the average variance of the compositional model shrinks much faster than the PNORMAL-TS. Because the exploration-exploitation of the Thompson sampling depends on the posterior uncertainty, the fast shrinkage in the posterior variance may indicate the under exploration of the model. This is predictable to a certain extent in the sense that one new triple with the compositional models induces multiple new compositional triples, so the uncertainties of entities and relations are measured less than those with non-compositional model. Most active learning algorithms utilise model uncertainty, and hence a model with augmented structures such as the relation compositions should be more careful about reflecting its uncertainty correctly.

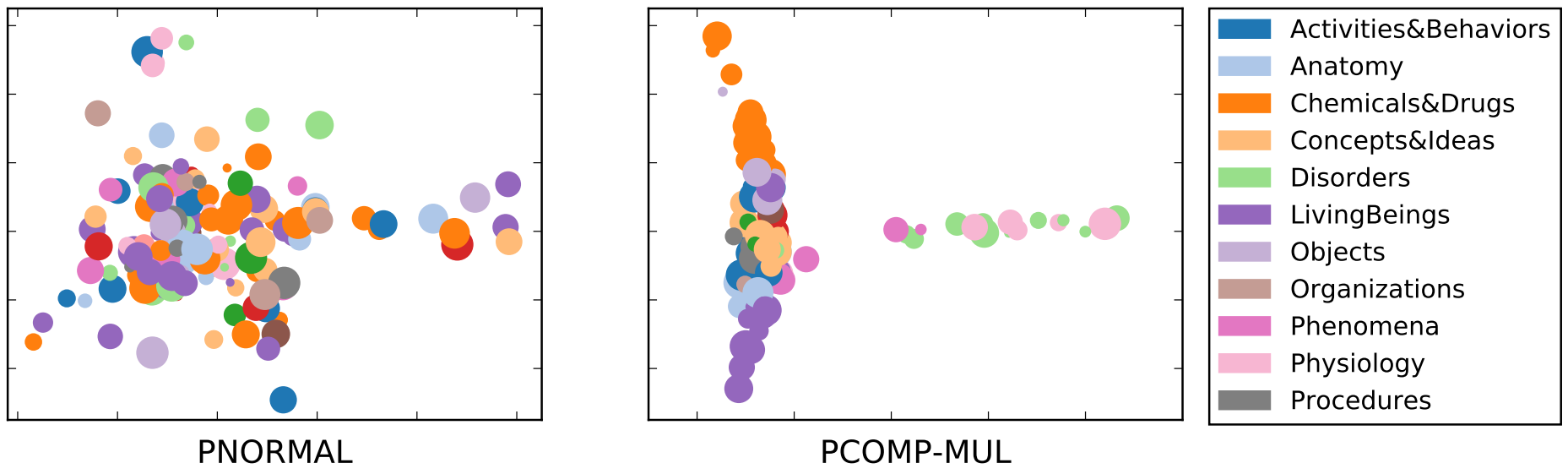

**Figure 4: Embedding learned entities of the UMLS dataset into a two-dimensional space through the spectral clustering. Entities with the same type are represented by the same color.**

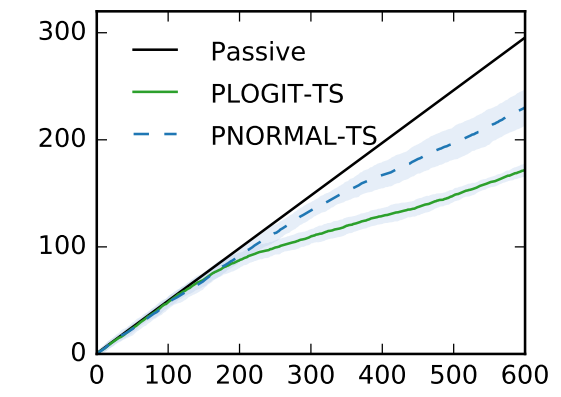

(a) Logistic: N=10, K=10, D=5

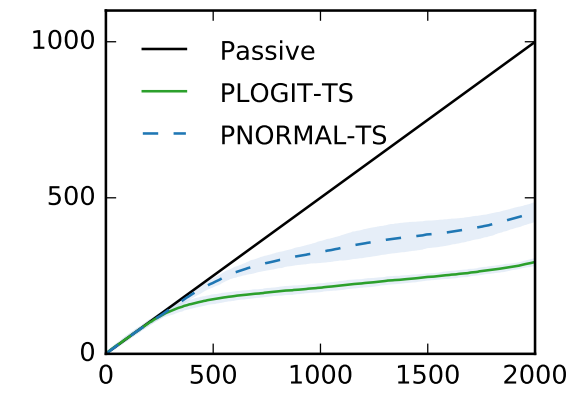

(b) Logistic: N=20, K=10, D=5

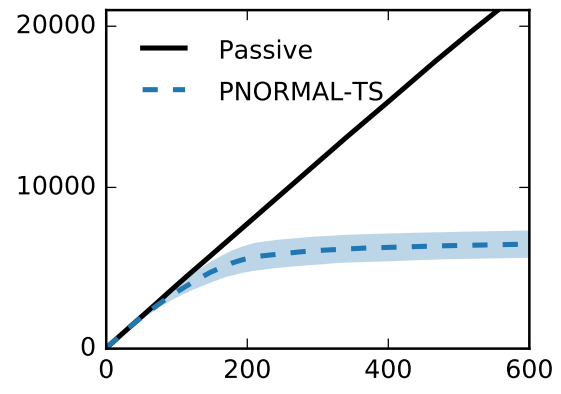

(c) Gaussian: N=10, K=10, D=5

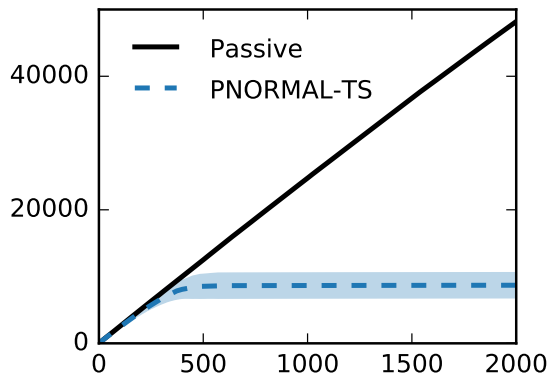

(d) Gaussian: N=20, K=10, D=5

**Figure 5: Cumulative regret of particle Thompson sampling with Gaussian and logistic output (PNORMAL-TS, PLOGIT-TS) against Passive learning on synthetic datasets with logistic (top row, a, b) and Gaussian (bottom row, c, d) output variables. The averaged cumulative regrets over 10 runs are plotted with one standard error. As the model obtained more and more labeled samples from Thompson sampling, the cumulative regrets increase sub-linearly.**

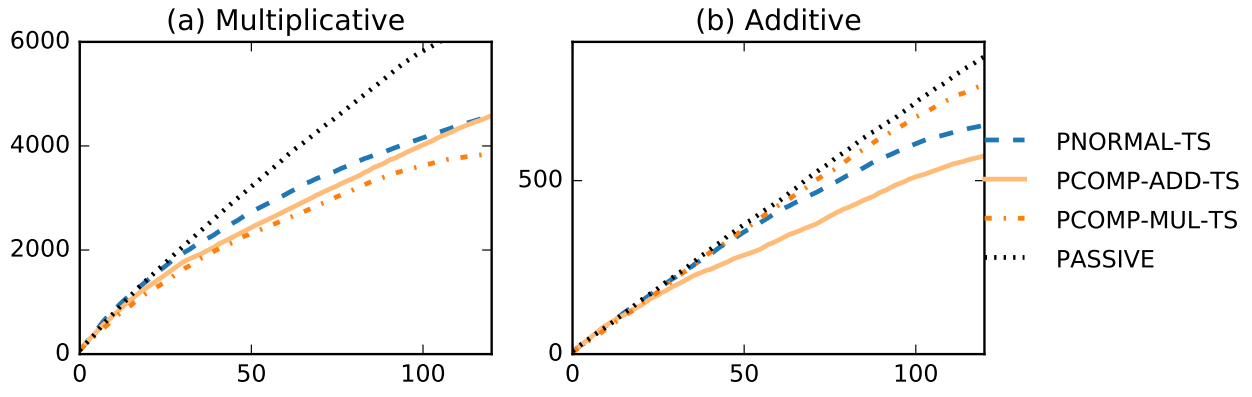

**Figure 6: Cumulative regret of particle Thompson sampling of the compositional models on synthetic dataset with N=5, D=5. The synthetic dataset has three relations (K=3); the first two are independently generated, and the third relation is composed by the first two relations. The dataset used in (a) is generated by the multiplicative assumption, and the dataset used in (b) is generated by the additive assumption.**

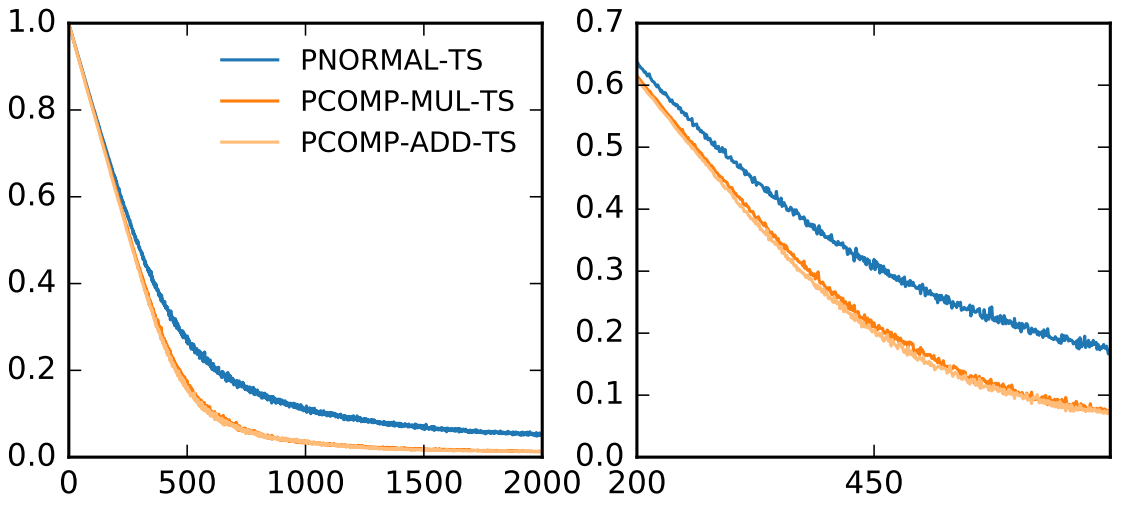

**Figure 7: Trace plot of mean posterior variance of the non-compositional model and compositional models. Y-axis denotes the average posterior covariance, and X-axis denotes the number of queries. The second plot magnifies the first plot.**